\documentclass[journal]{IEEEtran}

\usepackage{cite}

\ifCLASSINFOpdf
  \usepackage[pdftex]{graphicx}
  \graphicspath{{images/}}
  \DeclareGraphicsExtensions{.pdf,.jpeg,.png}
\else
\fi

\usepackage{amsmath,amsfonts,amssymb,bm}
\usepackage{mathrsfs}
\usepackage{algorithmic}
\usepackage{array}

\usepackage{xspace}

\ifCLASSOPTIONcompsoc
  \usepackage[caption=false,font=normalsize,labelfont=sf,textfont=sf]{subfig}
\else
  \usepackage[caption=false,font=footnotesize]{subfig}
\fi

\usepackage{url}

\usepackage{tabularx,booktabs,multicol,multirow} 
\usepackage{threeparttable}
\usepackage{algorithm}
\usepackage[shadow, textsize=footnotesize]{todonotes}
\usepackage[acronym, nonumberlist, nomain]{glossaries}
\usepackage{xcolor}

\glsdisablehyper
\makeglossaries

\newacronym{ml}{ML}{Machine Learning}
\newacronym{nn}{NN}{Neural Network}
\newacronym{dl}{DL}{Deep Learning}
\newacronym{ad}{AD}{Anomaly Detection}
\newacronym{dt}{DT}{Digital Twin}
\newacronym{knn}{KNN}{$k$-Nearest Neighbors}
\newacronym{ocsvm}{OCSVM}{One-Class Support Vector Machine}
\newacronym{pca}{PCA}{Principal Component Analysis}
\newacronym{lof}{LOF}{Local Outlier Factor}
\newacronym{if}{IF}{Isolation Forest}
\newacronym{cc}{CC}{Cluster Centers}
\newacronym{ae}{AE}{Autoencoder}
\newacronym{ffae}{FF-AE}{Feed Forward Autoencoder}
\newacronym{cnn}{CNN}{Convolutional Neural Network}
\newacronym{gpu}{GPU}{Graphics Processing Unit}
\newacronym{dnn}{DNN}{Deep Neural Network}
\newacronym{relu}{ReLU}{Rectified Linear Unit}
\newacronym{sae}{SAE}{Siamese Autoencoder}
\newacronym{cnnsae}{CNN-SAE}{Convolutional Siamese Autoencoder}
\newacronym{chp}{CHP}{Combined Heat and Power}
\newacronym{hmm}{HMM}{Hidden Markow Model}
\newacronym{svm}{SVM}{Support Vector Machine}
\newacronym{mlp}{MLP}{Multi Layer Perceptron}
\newacronym{mse}{MSE}{Mean Square Error}
\newacronym{ai}{AI}{Artificial Intelligence}
\newacronym{cnnae}{CNN-AE}{Convolutional Autoencoder}
\newacronym{snn}{SNN}{Siamese Neural Network}
\newacronym{tpr}{TPR}{True Positive Rate}
\newacronym{fpr}{FPR}{False Positive Rate}
\newacronym{roc}{ROC}{Receiver Operating Characteristic Curve}
\newacronym{auc}{AUC}{Area Under Curve}
\newacronym{hri}{HRI}{Honda Research Institute}
\newacronym{gmmhmm}{GMHMM}{Hidden Markow Model with Gaussian Mixture}
\newacronym{mae}{MAE}{Mean Absolute Error}
\newacronym{oop}{OOP}{Object Oriented Programming}
\newacronym{svc}{SVC}{Support Vector Classifier}

\usepackage{soul}

\newcommand{\new}[1]{ #1 }

\newcommand{\note}[1]{\textbf{\textit{{\color{blue}#1}}}\\}
\newcommand{\norm}[1]{\left\lVert#1\right\rVert}

\begin{document}
%

\title{Real-World Anomaly Detection by using Digital Twin Systems and Weakly-Supervised Learning}

\author{Castellani~Andrea,
        Schmitt~Sebastian,
        and~Stefano~Squartini,~\IEEEmembership{Senior~Member,~IEEE}
\thanks{A. Castellani and S. Schmitt are with the Honda Research Institute Europe GmbH, 63071 Offenbach, Germany (e-mail: acastellani@techfak.uni-bielefeld.de, sebastian.schmitt@honda-ri.de)}
\thanks{S. Squartini is with the Department of Information Engineering, Universit\`a Politecnica delle Marche, 60131 Ancona, Italy (e-mail: s.squartini@univpm.it)}
\thanks{Manuscript received XXXXXXX XX, 2020; revised XXXXXXX XX, 2020. DOI 10.1109/TII.2020.3019788}}

\markboth{IEEE Transactions on Industrial Informatics,~Vol.~XX, No.~X, XXXXXXX~2020, DOI 10.1109/TII.2020.3019788}%
{AUTHOR SURNAME \MakeLowercase{\textit{et al.}}: Real-World Anomaly Detection by using Deep Learning Algorithms and Digital Twin Systems}

\maketitle

\begin{abstract}
The continuously growing amount of monitored data in the Industry 4.0 context requires strong and reliable anomaly detection techniques. The advancement of Digital Twin technologies allows for realistic simulations of complex machinery\new{,} therefore\new{, it} is ideally suited to generate \new{synthetic} datasets for the use in anomaly detection approaches\new{,} when compared to actual measurement data. 
In this paper, we present novel \new{weakly-supervised approaches to anomaly detection for industrial settings, which make use of the Digital Twin to generate a training dataset which simulates the normal operation of the machinery, along with a small set of labeled anomalous measurement from the real machinery.
In particular, we introduce a clustering-based approach, called Cluster Centers (CC), and a neural architecture based on the Siamese Autoencoders (SAE), which are tailored for weakly-supervised settings with very few labeled data samples.
The performance of the proposed methods is compared against various state-of-the-art anomaly detection algorithms on an application to a real-world dataset from a facility monitoring system, by using a multitude of performance measures.
Also, the influence of hyper-parameters related to feature extraction and network architecture is investigated. }
We find that the proposed \new{SAE based solutions outperform} state-of-the-art anomaly detection approaches very robustly for many different hyper-parameter settings on all performance measures. 
\end{abstract}


\ifCLASSOPTIONpeerreview
 \begin{center} \bfseries EDICS Category: 3-BBND \end{center}
\fi
%
\IEEEpeerreviewmaketitle

\vspace{-0.2cm}
\section{Introduction}
%
%
%
%


\IEEEPARstart{I}{}n recent years, there has been a strong tendency to equip technical machinery, ranging from single machines to complete buildings and manufacturing plants with sensors to constantly monitor their operation, especially in the context of Industry 4.0 strategies\cite{preuveneers2017intelligent}.
Generally, component failures or complete system failures need to be avoided as this severely impacts the functioning of the machines and leads to significant increase in maintenance, overhaul and repair (MRO) costs.
In many situations, precursors of component failures can be observed in the time-series of measured sensor data and predictive maintenance approaches try to use this to reduce MRO downtime and cost\cite{CARVALHO2019106024}. 
An essential part of these approaches are robust and reliable anomaly detection methods which work in real-world settings.

Anomaly detection \cite{Chandola2009} refers to the problem of finding patterns in data that do not conform to expected or normal behavior.
It is an active area of research with a wide range of application areas, such as energy \cite{8859386}, manufacturing \cite{Chang2014}, network sensors \cite{8896029}, health care and video surveillance \cite{Nawaratne2020}. 
Anomaly detection techniques based on machine learning can be separated into different types of approaches \cite{Chalapathy2019}: supervised approaches, where a sufficiently large set of training samples with labelled data is available; unsupervised approaches, where only the unlabelled measurement data is available; and weakly-supervised approaches, where a large amount of unlabelled data with a very small set of labeled data is available. The distinction between these cases is not clear-cut, as there could be a supervised situation where a large amount of labelled  data is available, but this data is only from the normal operation. In that case, the training set does not reflect the true distribution of data in the real-world, and even more importantly, the most important information carrying data, i.e.\ labelled anomaly samples, are missing. 

For a practical application in realistic circumstances, \new{even if there are many data available,} only unlabelled data is usually available, since it requires a tremendous effort by human experts to manually create a fully labeled dataset with a large portion of the possible anomalous scenarios.
This requires unsupervised anomaly detection methods to be used. In those methods, the bulk statistics of the data is learned and the prevailing features of the data are considered normal. The decision whether a data sample is normal or not is then based on the comparison to the learned statistics so that  infrequent  data samples are more likely to be considered anomalous. 
This implies a very high class imbalance for the normal and the abnormal classes.  
Tuning the sensitivity of the anomaly detection algorithm is a rather difficult problem in such a situation. Even a very small false positive rate leads to a very large absolute  amount of wrongly detected anomalies which renders many approaches useless for practical applications.
Additionally, for most complex machinery, the definition of an anomaly is not so clear. There might exist normal operation modes, which are very rare and therefore, from a statistical viewpoint, anomalies. Also real-world machinery exhibits drift of data due to de-calibration of sensors, wear and tear and degradation of the machinery. So the performance of an anomaly detection which was trained at a specific instance in time might degrade over time due to the drift in the data
and regular re-training might be necessary.


In recent years, simulation technology has advanced substantially and even complex machinery can nowadays be simulated quite accurately. It is even possible to incorporate input data from actual measurements and to mirror the operational history of individual machines, which is known as Digital Twin systems first introduced in 2003 \cite{digitaltwinwhitepaper} (see also \cite{Tao2019}).
Depending on the effort spent to create the Digital Twin, it can be realistic simulation which captures the qualitatively correct behavior of the machinery, or an almost perfect digital copy whose output can be directly compared to the real-world machinery.
The former can be at least used to create a large dataset containing data samples of normal operation conditions which can be utilized in machine learning approaches.   
The latter can play a key role in the anomaly detection problem \cite{Sugumar2019,Balta2019, Pileggi2019}, when it runs in parallel to the physical system with the same input values and environmental conditions. 

In this paper, we present novel weakly-supervised approaches to anomaly detection which we apply to the data from a company facility monitoring system. We generate a large dataset of normal operation data covering a complete year of operation using a Digital Twin simulation of the system.  This enables the accurate unsupervised learning of the statistics of the normal operation states statistics including the very rare but normal states. In order to increase the sensitivity for anomalies, we additionally use a very small dataset of  labeled anomalous samples which is obtained from the real-world measurements. This weakly-supervised approach has the advantage to produce a substantially lower false positive rate making it suitable for usage in the real-world. Additionally, the Siamese Networks can be retrained very efficiently as soon as new labelled data is available, making it ideal for refinement during usage as well as for adjusting to drift  and re-configurations of the monitored machinery during operation.

The main contributions of this work are: (i) 
	 novel approaches to anomaly detection using data from a Digital Twin simulation for the normal operational state of a machinery,
	(ii)  a clustering-based algorithm capable to solve the anomaly detection task in both unsupervised and weakly-supervised settings,
	(iii) a Siamese Autoencoder architecture for weakly-supervised anomaly detection where very few labelled training samples are needed to improve the performance over unsupervised anomaly detection methods,
	(iv) a thorough comparison of experimental results of the proposed methods and their performance to state-of-the-art algorithms for anomaly detection.

The rest of the paper is organized as follows.
Section~\ref{sec:related} presents an overview of the recent literature on anomaly detection on time-series and the usage of Digital Twin.
Section~\ref{sec:proposed} shows the proposed approach and presents the developed algorithms.
Section~\ref{sec:DigitalTwin} presents the used Digital Twin framework.
Section~\ref{sec:comparative} shows the comparative anomaly detection algorithm investigated in this work and Section~\ref{sec:Setup} presents the dataset and describes the experimental setup.
Section~\ref{sec:results} discusses the results obtained in our experimentation.
Finally, Section~\ref{sec:conclusion} concludes the paper and presents future developments.


\section{Related Works}\label{sec:related}
Anomaly detection can be approached in different ways depending on the kind of data available and the requirements of the particular use case.
Chandola et al. \cite{Chandola2009} presented a survey of different anomaly detection techniques applied to different domains and recently the problem has been addressed from the perspective of Deep Learning by Chalapathy et al. \cite{Chalapathy2019}.
One category of deep anomaly detection methods is based on reducing the dimension of the input by mapping it to a low dimensional manifold followed by reconstruction.
These networks are usually trained in unsupervised fashion with a dataset containing only data that reflects the normal state of operation.
The reconstruction error is then used to distinguish anomalies with several different neural architectures, for example  Autoencoders (AE) \cite{Principi2019}, Long Short Memory Network (LSTM) \cite{Wu2019a}, Adversarial Autoencoder (AAE) \cite{9016153} or Generative Adversarial Networks (GAN) \cite{Jiang2019}.
When dealing with multivariate time-series data, Canizo et al. \cite{Canizo2019} use a CNN-RNN capable to reach an AP score of 0.994 on a real-world dataset, but the proposed architecture is trained with a fully supervised approach.
Other approaches to anomaly detection make use of data-driven statistical methods \cite{Wang2019, 8859386}, unsupervised clustering \cite{Li2019} and density-based clustering \cite{Chang2014}.

Tao et al. \cite{Tao2019} presents a detailed survey of the actual state-of-the-art of the Digital Twin in industry. Several applications are presented in different areas of design, production, prognostic and health management. 
To the best of our knowledge, there are just few references in literature to the usage of Digital Twin in the context of anomaly detection \cite{Balta2019, Pileggi2019, Sugumar2019} and all of those are more focused to present the Digital Twin system rather than a anomaly detection algorithm.

While there are several supervised and unsupervised approaches to anomaly detection described in literature, only few weakly-supervised approaches have been proposed.
In \cite{8758199} the authors use domain adversarial training to transfer the knowledge learned from a normal dataset to another with few labeled samples.
The Siamese Networks approach was introduced for signature verification in \cite{bromley1994signature} in a fully supervised setting.
A first weakly-supervised approach using Siamese Networks was made by Koch et al. \cite{koch2015siamese}.
Regarding anomaly detection, there are some applications of Siamese Autoencoders including  human fall \cite{Droghini2019} and medical application \cite{Alaverdyan2020}. Both architectures make use of Siamese Autoencoder to extract features from the data, but the anomaly detection process is made by an additional classifier, a KNN in the former and OCSVM in the latter.
Utkin et al. \cite{Utkin2017} presents a variation of the Siamese Autoencoder with application to robotics.
To the best of our knowledge, no previous attempts to approach the anomaly detection problem with weakly-supervised Siamese Networks and the combination of Digital Twin and real-world data have been proposed.


\section{Proposed Algorithms}\label{sec:proposed}

\begin{figure}[!t]
	\centering
	\includegraphics[width=\linewidth]{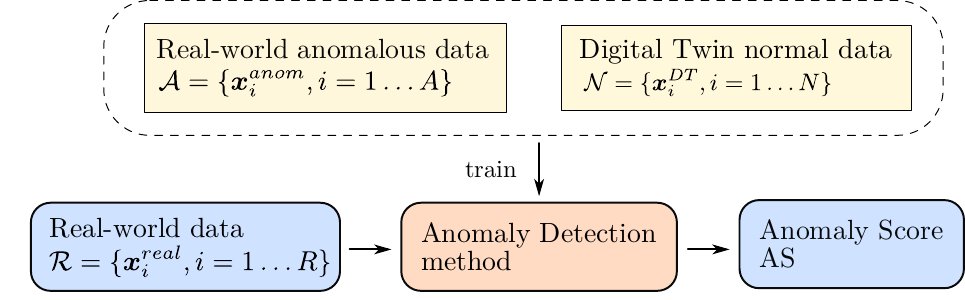}
	\caption{Proposed pipeline for Anomaly Detection.}
	\label{fig:pipeline}
\end{figure}

The overall pipeline of the proposed anomaly detection scheme is sketched in \figurename~\ref{fig:pipeline}.
It consists of a training phase of an anomaly detection algorithm (dotted rectangle) with the combination of two datasets: normal operation dataset $\mathcal{N}$, generated with the Digital Twin simulation, and a small set  $\mathcal{A}$ of labeled anomalous samples from the real-world measurement system.
During the regular operational phase (blue boxes), the physical values from the machinery provide a real-world dataset $\mathcal{R}$, which is fed to the trained anomaly detection algorithm. An Anomaly Score (AS) is calculated for each data sample and it is considered an anomaly if its score exceeds a certain threshold.

\subsection{Cluster Centers}
We propose the \acrfull{cc} algorithm as a weakly supervised classification approach to the anomaly detection problem.
In a first step, the \acrlong{dt} dataset of normal samples $\mathcal{N}$ is processed with an unsupervised clustering algorithm, which identifies distinct normal operational modes of the industrial machinery. Here, we use a $k$-means approach, but in general any clustering algorithm is suitable, 
as long as it is possible to calculate the set of cluster centers, $\mathcal{C}=\{\bm{c}_j\: \mathrm{for}\: j=1,...,N_C\}$.  

The number of clusters $N_C$ and the clusters themselves do not need to represent the real semantically correct operational modes of the machinery, but rather organize the data into different clusters which cover  the complete statistical variations of data found during normal operation.  For this reason, the number of clusters should be considerably larger than the (guessed or known) true number of operational states so the clusters can also cover larger variations,  transitions and switching behavior properly.

The cluster centers $\bm{c}_j$ from the clustering algorithms \new{of only normal data samples from the Digital Twin} can already be used to predict the operational clusters on the  real-world data set $\mathcal{R}$.
\new{The distribution of cluster centers reflects only normal operation, and only from the Digital Twin. The expectation is that the normal operations data from the real-world aligns with this clustering, while anomalous data samples are quite distinct from normal operation data and cannot be assigned reasonably
to these cluster centers. Therefore, the cluster centers can} provide a fully unsupervised estimate for the AS, which is given by the distance ($d$) from each data point ($\bm{x}_r$) to its nearest cluster center ($\bm{c}$). For the $k$-means approach we use the Euclidean distance, but any other distance measure could also be used.

The set of labeled anomalies, $\mathcal{A}$, is used to refine the AS of each real-world sample $\bm{x}_r$ by adding a penalty term,
\begin{equation}\label{eq:CC}
AS(\bm{x}_r) = {\underbrace{
    \vphantom{\frac{1}{\min(d(\bm{x}_r, \bm{x}_a) + \zeta)}}
        \min_j(d(\bm{c}_j, \bm{x}_r))}_{\text{Unsupervised AS}}} + {\underbrace{ \eta\, \frac{1}{\min_a(d(\bm{x}_r, \bm{x}_a)) + \zeta}}_{\text{Penalty term}}},
\end{equation}
with the cluster centers $\bm{c}_j \in \mathcal{C}$, the anomalous data samples $\bm{x}_a \in \mathcal{A}$,
a hyper-parameter $\eta >0$ determining the influence of the penalty
and a regularization factor $\zeta>0$. This penalty term increases the AS whenever a known anomaly is close to the data sample. \new{With this, the information about the known anomalies is used to refine the information provided by the cluster centers, which represents only normal operation, and allows for a better discrimination between normal an anomalous data samples.}

The proposed \acrshort{cc} and a regular \acrfull{knn} clustering approach   differ in the way the AS is calculated. We use the distance to the nearest cluster center, whereas in a KNN approach the nearest data samples are used. 
Also, the penalty term is not used in regular KNN approaches.


In the above algorithm more hyper-parameters could have been introduced, for example,  an exponent different from 1 for the penalty term in~\eqref{eq:CC} which would allow for  asymmetric weighting for small and large distances. This was purposely not done to keep the approach as simple as possible. Also, instead of including the labeled anomaly samples via a penalty term after the clustering, they could have been included from the beginning by using a constrained clustering algorithm. This was not done in order to deal more easily with a constantly growing and changing set of labeled anomalies, which can be used in the AS without redoing the clustering.

\subsection{Siamese Autoencoder}
The Siamese Networks consist of two identical networks with shared weights which can be efficiently used to decide if a pair of input data samples comes from the same distribution or not.

\begin{figure}[!t]
	\centering
	\includegraphics[width=0.7\linewidth]{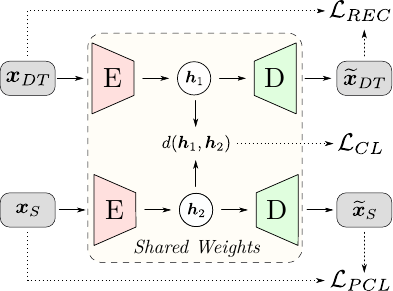}
	\caption{Siamese Autoencoder structure with the three loss functions.}
	\label{fig:Siamese_high_level_color_V2}
\end{figure}

The structure of the proposed architecture is shown in \figurename~\ref{fig:Siamese_high_level_color_V2}.
It comprises an encoder $E$ which maps the input data sample $\bm{x} \in \mathbb{R}^N$ to a latent representation $\bm{h}(\bm x)\in\mathbb{R}^M$, with $M < N$. The decoder network $D$ takes the latent representation and reconstructs a data sample in the original space,  $\tilde{\bm{x}}\in \mathbb{R}^N$.
\new{We refer the reconstruction of a given input sample, $\bm{x}$, as $\tilde{\bm{x}} = D(E(\bm{x}))=D(\bm{h}(\bm x))$, and its the latent representation as $\bm{h}(\bm x)=E(\bm{x})$.}

The main goal of the proposed network is to reconstruct normal data samples with low error and to create a clear separation of the normal and the anomalous data distribution in the latent space. This can be achieved by the following behaviour of the network: 
\begin{enumerate}
\item Normal data samples should be reconstructed as good as possible: $\bm{x} \, \new{\simeq} \, \tilde{\bm{x}}$ for $\bm{x} \:\in\: \mathcal N$. 
\item The distance between two latent representations of any pair of normal data samples should be as small as possible: $d(\bm h(\bm{x}),\bm h(\bm x'))$ small for $\bm{x},\bm x' \:\in\: \mathcal N$. 
\item Anomalous data samples should not be reconstructed well, i.e.\ their reconstruction error should be much larger \new{than} the reconstruction errors for normal data: $d(\bm{x},\tilde{ \bm x})$ large for $\bm x \:\in\: \mathcal A$.
\item The distance between the latent representations of a normal and an anomalous data sample should be large: $d(\bm h(\bm{x}),\bm h(\bm x'))$ large for $\bm{x} \:\in\: \mathcal N$ and  $\bm{x}' \:\in\: \mathcal A$.
\end{enumerate}

The desired behaviour can be achieved by implementing an appropriate training procedure for the neural network.
The network always evaluates a pair of data samples, $(\bm x_{DT}, \bm{x}_S)$,  where  the first sample is always taken from the normal data set, $\bm x_{DT} \:\in\: \mathcal N$. In our approach normal data samples for training are always taken from the Digital Twin dataset, which is why we include the subscript $DT$ for clarity.  The second sample can be either from the normal or the anomalous dataset,  $\bm x_{S} \:\in\: \mathcal N \cup \mathcal A$.
The loss function for training then has the following three contributions 
\begin{equation}
    \mathcal L =\mathcal{L}_{REC} + \mathcal{L}_{CL}+\mathcal{L}_{PCL}
\end{equation}
where each contribution is calculated as sums over all pairs of data $(\bm x_{DT_i},\bm{x}_{S_i})$ and are  given by the following expressions: 
\begin{itemize}
	\item \textbf{Reconstruction Loss:} the \acrfull{mse} between the input  and its reconstruction for normal operation data, which is just the cost function of a typical Autoencoder:
	\begin{equation}\label{eq:rec}
	\mathcal{L}_{REC} = \frac{1}{N} \sum _{\mathrm{pairs} \:i} \big ( \new{\tilde{\bm{x}}_{DT_i}} - \bm{x}_{DT_i} \big )^2 .
	\end{equation}
	
	\item \textbf{Contrastive Loss:} the Euclidean distance between the latent vectors of the input pair with a local modification:
	\begin{multline}\label{eq:cl}
	\mathcal{L}_{CL} =\frac{1}{2N} \sum _{\mathrm{pairs} \:i} \Bigg(
	(1 - Y_i)\, d\left ( E(\bm{x}_{DT_i}),E( \bm{x}_{S_i} )\right )^2 \\
	+ Y_i \left \{ \max\big ( 0, m-d\left (E( \bm{x}_{DT_i}),E( \bm{x}_{S_i}) \right ) \big ) \right \}^2\Bigg),
	\end{multline}
	 where $Y_i=0$ if $\bm{x}_{S_i} \in \mathcal{N}$ and   $Y_i=1$ if $\bm{x}_{S_i} \in \mathcal{A}$.
	 This contribution minimizes the difference for input from the same class ($Y=0$) while maximize the differences for input of different class ($Y=1$). The parameter $m > 0$ allows only samples whose distance is less than the radius defined by $m$ to contribute to the loss function in order to avoid the domination of individual samples.
	\item \textbf{Partial Contrastive Loss:} enforces a large reconstruction error for anomalous data sample, which can also be viewed as a gradient inversion in the Autoencoder training process:
	\begin{equation}\label{eq:pcl}
	\mathcal{L}_{PCL}=\frac{1}{2N} \sum _{\mathrm{pairs} \:i} Y_i \:\max\big ( 0, m-d\left (E( \bm{x}_{DT_i}),E( \bm{x}_{S_i}) \right ) \big ) 
	\end{equation}%
\end{itemize}



Due to the possibility to construct a very large training dataset of distinct pairs from the large normal dataset and very small anomalous dataset, 
this approach is able to deal with extremely unbalanced datasets, $|\mathcal N|\gg|\mathcal A|$.

After training, the AS for a new real-world data sample $\bm{x}_r \in \mathcal{R}$ is calculated as:
\begin{equation}
AS(\bm x_r)\!=\!{\underbrace{\vphantom{\frac{1}{N'} \sum_{i=1}^{N'}}\left(\new{\tilde{\bm{x}}_{r}}  - \bm{x}_r\right)^2}_{\text{Reconstruction error}}} + {\underbrace{\frac{1}{N'}\!\sum_{i=1}^{N'} \norm{E(\bm{x}_{DT_i}) - E(\bm{x}_r)}_2}_{\text{Embedding distance}}},
\end{equation}
where $\bm{x}_{DT_i}$ are from a subset of the normal operation dataset from the training phase  with $N'\leq N$ elements.

In this work we employ the Siamese architecture for two types of input data. In the so-called \acrfull{sae} we extract feature vectors from the measurement time series data and use standard feed forward autoencoder architectures for encoder and decoder which are chosen to be symmetric.    
The second variant called \acrfull{cnnsae} directly operates on raw time series data and employs 1-D Convolutions, where  encoder and decoder are also chosen to be symmetric.


\section{Digital Twin}\label{sec:DigitalTwin}
The \acrlong{dt} model used in this work simulates the electrical power system, the Heating, Ventilation and Air Conditioning (HVAC), Combined Heat and Power (CHP) systems for a medium sized company facility. The simulation is realized with the \textit{Green City library}\footnote{\url{http://ea-energie.de/en/products/green-city-simulationsbibliothek-2-2/}} and the  \textit{SimulationX} software based on Modelica programming language.
The detailed process of the calibration of the \acrlong{dt} with the measurement data from the machinery is presented in a separate publication \cite{pub4012}.
This work focuses on the \acrshort{chp} model which takes the actual measurements  form weather data and total power demand time-series as input.

Due the high complexity of modelling realistic  machines and the heating and cooling of a company facility, the Digital Twin is only able to simulate with very high fidelity power and energy consumption of the machines in a reasonably large window of time, i.e.\ not less than few hours. The exact transient behavior of the machines cannot be simulated exactly so comparing raw time series between \acrlong{dt} and the measurement data is not suitable on the sampling rate of one data point per minute. However, we have validated the Digital Twin model with several periods of correct operation of the machine in 2018 and there is a discrepancy with the real-world electrical and thermal energy production of $2.18\%$.


\section{Comparative Methods}\label{sec:comparative}

The simplest method for deriving an AS is given by directly comparing the raw time-series of the measurement with the simulation and taking the \acrfull{mae}. 
As state-of-the-art  algorithms for anomaly detection we employ an Isolation Forest (IF), which builds an ensemble of isolation trees for a given dataset, where anomalies are those data samples with short average path lengths in the trees.
In the \acrfull{knn} approach the measure of outlierness of an observation is obtained based on the distance to its neighbors. 
We also compare the results of supervised Support Vector Machine (SVM), its unsupervised variant One-Class SVM (OCSVM), as well as to the \acrfull{lof}, which measures the outlierness of data points by its local deviation of densities with respect to its neighbors.
The unsupervised dimensional reduction method \acrfull{pca} is used as an anomaly detection technique by taking the reconstruction error of each sample as AS.
Finally we also compare to a Deep Learning-based \acrfull{ffae}  and a supervised Multi Layer Perceptron (MLP).

All unsupervised algorithms are trained on the Digital Twin dataset $\mathcal N$ and then evaluated against the real-world data $ \mathcal R$. 
In addition, one OCSVM is also trained and tested on real-world data only. The weakly-supervised and supervised algorithms are trained with the Digital Twin dataset and few randomly sampled anomalies from $\mathcal A$.


\section{Experimental Setup}\label{sec:Setup}

\subsection{Dataset Description}
In this study, we use a dataset from a medium sized company, where an infrastructure monitoring system recording various energy-related modules is installed.
The recorded data consists of  a large number of sensors for heat, cold and electricity consumption and production, as well as a local weather station to monitor the ambient environment.
This work focuses on data recorded from the CHP module, where natural gas is burned to produce heat and electrical power. The recorded time-series are: consumed and produced energy, produced heat, heating fluid volume flow, and flow and return temperatures for the heating fluid, as shown in Table~\ref{tab:data_hypepar}. 

The dataset consists of data from \textit{November 2017} to \textit{February 2019}.
The sampling rate is one data point per minute, leading to a total of 658081 sample points.

The synthetic Digital Twin dataset consists of the same time-series and the same processing as for the real-world dataset, but only the year 2018 was simulated. 

The anomalous real-world dataset has been labeled manually. A total of 100 failure instances have been identified  which translates to 24.2\%  of the total number of samples being  anomalous. 
There are two kinds of anomalies. One is produced by sensor failures which result in a flat time series, and which are rather easy to detect. More interesting and harder to detect are those situations in which each time-series appears to provide a valid measurement, but the combination multiple sensor reading does not reflect normal operation. For the CHP, this is the case when, for example, the ambient temperature is rather low but still no electricity and heat are produced. 
\new{An example of a normal and anomalous data sample is shown in Fig.~\ref{fig:cnn_reconstruct}.}

\subsection{Data Pre-Processing and Evaluation Metrics}

\begin{table}[!t]
	\centering
	\caption{Data Hyperparameters explored in the \new{partial} grid search. Values in bold have been used for the final configuration.}
	\label{tab:data_hypepar}
	\begin{tabular*}{\linewidth}{l @{\extracolsep{\fill}} l}
		\toprule
		\textbf{Hyperparameter} & \textbf{Value} \\
		\midrule
		\multirow{9}{*}{Variables} & \textbf{Ambient Temperature} ($T\text{a}$) \\
		& \textbf{Produced Thermal Power} ($P_{\text{Th}}$)\\
		& Produced Thermal Energy ($E_{\text{Th}}$)\\
		& \textbf{Produced Electrical Power} ($P_{\text{El}}$) \\
		& Produced Electrical Energy ($E_{\text{El}}$)\\
		& Water flux ($Flux$)\\
		& Water flow temperature ($T_{\text{Flow}}$)\\
		& Water return temperature ($T_{\text{Return}}$)\\
		& Flow/return temperature difference ($T_{\text{Diff}}$)\\
		\midrule
		\multirow{6}{*}{Statistical Features} & \textbf{Mean} ($\mu$)\\
		& \textbf{Standard deviation} ($\sigma$)\\
		& Skewness \\ 
		& Kurtosis \\
		& Sum\\
		& Root Mean Square\\
		\midrule
		Time Features & True, \textbf{False}\\
		\midrule
		Contextual Feature & True, \textbf{False}\\
		\midrule
		Window Length & 240, 360, 480, 720, \textbf{1440} \\
		\midrule
		Window Stride & 30, \textbf{60}, 120\\
		\bottomrule
	\end{tabular*}
\end{table}

For all approaches except the \acrshort{cnnsae}, we use features derived from the raw data as input to the models. The implemented statistical features are listed in Table~\ref{tab:data_hypepar} and are extracted with a sliding window approach with parameters also shown in the aforementioned table. 
The contextual feature refers to the number of machine shutdowns (which can be easily detected from the data) and the total working time within the time window.
The time features include the information related to working days and season of the year.

The feature vectors are composed of data from different domains and scales, and therefore need to be standardized with the \textit{z-score} $:Z = (X - \bar{x})/\sigma$,
where $\bar{x}$ is the mean of  $X$ and $\sigma$ its standard deviation.
Regarding time and contextual values, One-Hot Encoding is used to represent them as linearly uncorrelated vectors.

We evaluate the performance of the discussed approaches by using multiple metrics.
We use the $F_2$ Score, the Area Under Receiver Operating Characteristic curve (AUC ROC) and the Average Precision (AP) as a measure of the Area Under the Precision-Recall Curve (PRC).
In the situation with highly imbalanced class sizes, it was shown that the PRC is more informative than ROC \cite{Saito2015}, since it better reflects the correct prediction of the minority class.

\subsection{Implementation and Training Details}\label{sec:impl}
The state-of-the-art anomaly detection algorithms used for the comparison are taken from the open source library \textit{PyOD} \cite{zhao2019pyod}.
The neural networks are implemented with the Keras library 
and \textit{TensorFlow 1.13} back-end.
The data processing and training of the  algorithms is  done on two workstations, one with a quad-core CPU (Intel Core i5-7300HQ) and 16 GB of DDR4 RAM and another one with an Intel Xeon X5680 (6 cores and 12 threads), 128 GB of DDR4 RAM and a NVIDIA TITAN X GPU.


\begin{table*}[!t]
	\centering
	\caption{Autoencoders Networks hyperparameters explored in the Grid Search.}
	\label{tab:HYPER}
	\begin{threeparttable}	
		\begin{tabular*}{\linewidth}{l c @{\extracolsep{\fill}} ccc }
			\toprule
			\textbf{Hyperparameter}& \textbf{Range} & \textbf{\acrshort{ffae}} & \textbf{\acrshort{sae}} & \textbf{\acrshort{cnnsae}}\\
			\midrule
			Batch size & $\{16, 32, 64, 128\}$ & 64 & 128 & 32 \\
			Learning rate & $\left[10^{-2}, 10^{-5}\right]$ & $10^{-4}$ & $10^{-3}$ & $10^{-4}$\\
			Dropout & $\left[0, 0.5\right]$ & 0 & 0 & 0.1 \\
			Epoch & $\left[50, 200\right]$ & 150 & 150 & 100 \\
			Activation function & $\{\text{tanh}, \text{ReLu}, \text{sigmoid}\}$ & $\text{tanh}$& $\text{ReLu}$ & $\text{ReLu}$\\
			\midrule
			Nr. of Encoder/Decoder layers & $\left[1, 5\right]$ & 3 & 4 & 5\\
			Encoder/Decoder neurons / kernel\tnote{$*$} & $\left[8, 512\right]$ & 64 32 16 & 128 64 32 16 & 32 64 128 128 256\\ 
			Encoder/Decoder kernel length & $\left[1,10\right]$ & / & / & 10 7 5 4 3\\
			Encoder/Decoder kernel stride & $\left[1,3\right]$ & / & / & 2 2 2 3 1\\
			Nr. of fully-connected layers & $\left[1,3\right]$ & / & / & 1\\	
			Fully-connected neurons & $\left[10, 100\right]$ & / & / & 60 \\
			Embedding space dimension & $\left[1,3\right]$ & 3 & 2 & 2\\
			\bottomrule
		\end{tabular*}

	\begin{tablenotes}\footnotesize
		\item[$*$] Only for Convolutional Networks
	\end{tablenotes}

\end{threeparttable}
\end{table*}

We use 10-fold Cross Validation, where the Digital Twin dataset and the subset of labeled anomalies for training are divided in 10 folds to train and validate the algorithm's performance.
The real-world dataset is divided in two parts, $20\%$ for validation  and $80\%$ for the final test evaluation.

\begin{figure}[!tb]
    \centering
    \includegraphics[width=0.5\linewidth]{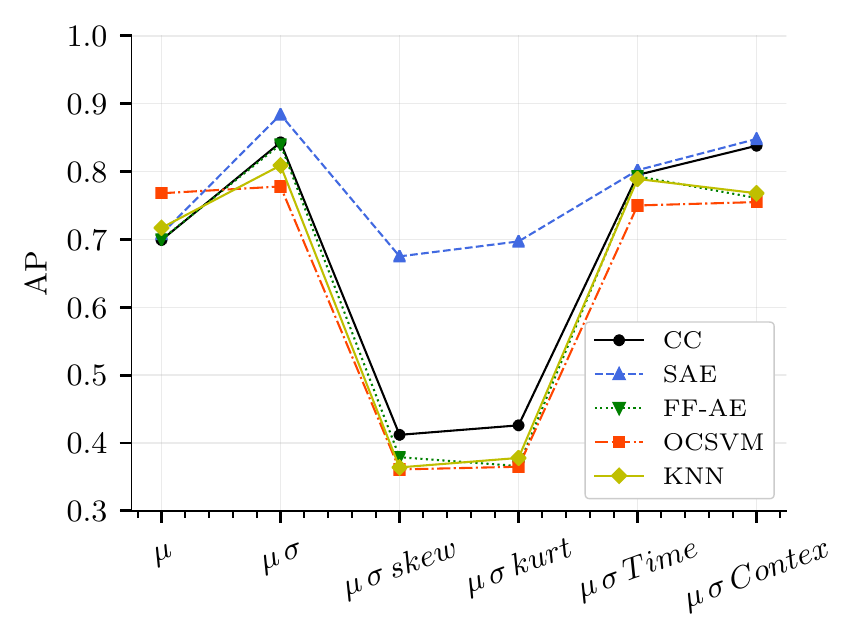}
    %
    %
    %
  \includegraphics[width=0.5\linewidth]{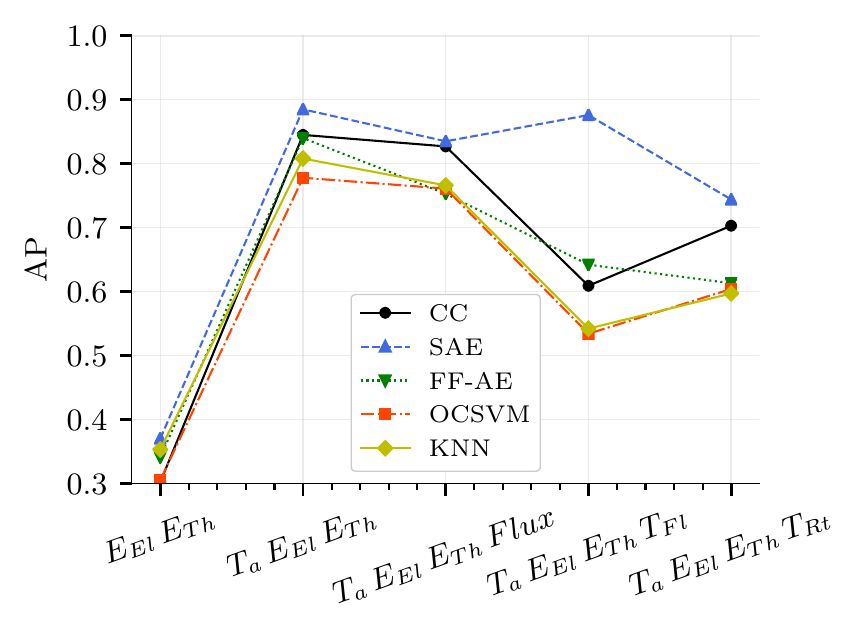}
    
    \caption{Comparison of the AP performance for different settings for the selected statistical features (left) and  input time series (right). The symbols and names shown on the $x$-axis are explained in Table~\ref{tab:data_hypepar}. 
    Other features are those shown in bold in Table~\ref{tab:data_hypepar}.}
    \label{fig:comp}
\end{figure}

\new{In order to determine the best performing combination of hyper-parameters for the feature extraction and the architectures of the Neural Networks, a Partial Grid Search \cite{bergstra2012random} has been used.
Instead of searching in the complete hyper-parameter space by computing all the possible combination of the parameters, the search is organized as an iterated line-search.  All parameters except one are fixed and the minimization is performed only along the dimension of the free parameter. Then, the best performing value for this parameter is selected and the search is done along another parameter dimension.  
The resulting hyper-parameters related to data and feature extraction are shown in Table~\ref{tab:data_hypepar}.
The values in bold constitute the best configuration of parameters, determined by the search approach with Cross Validation.
An example of this approach is reported in Fig.~\ref{fig:comp}, which illustrates the performance variations for some different settings related to the feature selection.}
The left panel shows the performance as function of the statistical quantities included in the feature vector. 
The performance for different input time-series is instead shown in the right panel. 
The qualitative changes and trends in the performance are the same for all algorithms, only the \acrshort{sae} sometimes has a slightly different trend. The variation of the statistical features (left panel) shows that the best performance is achieved when only mean ($\mu$) and standard deviation ($\sigma$) are used. 
The inclusion of higher statistical moments like skewness and kurtosis  leads to a degradation of the performance most likely due to the fact that the short time-scales, switching and state transitions, are not captured well by the Digital Twin simulation and therefore cannot contribute to a successful classification on the real-world data. By inspecting the curve in the right panel we can state that the ambient temperature $T_a$ is a rather important measurement, since not including it in the feature vector leads to a performance drop from around 0.8 to around 0.3. This was expected since the CHP is controlled on that variable. More sensors do not improve the performance, also the inclusion of the flow and return temperatures do not seem to be effective in this sense, for most algorithms. Only the \acrshort{sae} shows high performance when the flow temperature is used.

The resulting final dataset of feature vectors includes 8737 samples from the Digital Twin and 10945 samples  from  the  real-world, of which 2162 samples are labeled anomalies. Each  sample  consists of a feature vector of dimension 6.

The hyper-parameters of the Neural Network architectures are shown in Table~\ref{tab:HYPER}, with ranges used in the \new{search approach} reported in the first column. 
We use a symmetrical structure for each Autoencoder.
In particular, in the CNN case, the encoder is composed of five convolutional layers followed by a global max-pooling layer and then a fully-connected layer. The decoder has five de-convolutional layers (up-sampling layer followed by a convolutional layer).
In order to revert the loss of information in the global max-pooling layer, a skip-connection with linear activation is used between the last convolutional layer of the encoder and the first de-convolutional layer of the decoder. This is crucial to reconstruct the input data.

For training all neural architectures, we use the Adam optimizer with \textit{``LeCun Uniform"} weight initialization \cite{lecun2012efficient} and early stopping.
We also apply gradient clipping with maximum $L_2$ norm of 4 to improve training stability.
For weakly-supervised methods, the set of labeled anomalies for training is composed of only 10 samples, randomly selected from the available labeled data.
For the \acrshort{cc} algorithm, we set $\zeta = 0.001$ and $\eta = 0.15$, since those are the best performing combination.
For the OCSVM and SVC, we use a RBF kernel with $C = 1$. 
For the MLP we use a network with 3 hidden layer of 50 neurons each.
Each experiment has been repeated 10 times and, as results, we present the mean and standard deviation of the performance measures calculated on the test set.
\new{The AS threshold, $\theta_{AS}$ is the value of AS in correspondence of which the $25\%$ of the test-data has higher AS and it is considered anomalous.
This choice of $\theta_{AS}$ is based on an a-priori assumption of the expected anomaly rate of the machinery under investigation.
Furthermore, the $\theta_{AS}$ needs to calculate confusion matrix of the classifier, and then the $F_2$ score. In our experiments we use the AP score to compare the performance of different algorithm and for the hyperparameter search, because it is threshold-free and it is showed in \cite{Saito2015} that it is the best metric to use when dealing with an imbalanced classification task.}

\section{Experimental Results}\label{sec:results}

\begin{table*}[!tb]
	\centering
	\caption{Performance values of all evaluated algorithms\new{, using the Digital Twin to generate the training-set.} The globally best scores are highlighted in bold, while the best unsupervised scores are in italic font. For the weakly supervised methods only 10 additional labeled anomalies were used during training.}
	\label{tab:results}
	\begin{threeparttable}
		\begin{tabular*}{\linewidth}{l @{\extracolsep{\fill}} l l l l l}
			\toprule
			\textbf{Training Approach} & \textbf{Algorithm} & \textbf{Train Time} & \textbf{AP} & \textbf{AUC ROC} & $\bm{F_2}$\\
			\midrule
			\multirow{8}{0.15\linewidth}{Unsupervised}
			&\textit{\acrfull{mae}} & 0.41 s & $0.692 \pm 0.002$ & $0.792 \pm 0.001$  & $0.512 \pm 0.001$ \\
			&\textit{\acrfull{if}} & 0.54 s & $0.603 \pm 0.024$  &  $0.888 \pm 0.011$ & $0.697 \pm 0.026$\\
			&\textit{\acrfull{ocsvm}}& 3.54 s & $0.778 \pm 0.001$  &  $0.843 \pm 0.001$ & $0.738 \pm 0.001$\\
			&\textit{\acrfull{knn}} & 0.56 s & $0.807 \pm 0.008$ & $0.893 \pm 0.002$ & $0.720 \pm 0.002$\\
			&\textit{\acrfull{pca}}  & 0.09 s& $0.519 \pm 0.001$ & $0.649 \pm 0.001$ & $0.478 \pm 0.001$\\
			&\textit{\acrfull{lof}} & 0.18 s& $0.811 \pm 0.002$ & $0.869 \pm 0.001$ & $0.711 \pm 0.002$\\
			&\textit{\acrfull{ffae}} & 32.4 s& $\mathit{0.842 \pm 0.007}$ & $\mathit{0.914 \pm 0.006}$ & $\mathit{0.746 \pm 0.013}$\\
			&\textit{\acrfull{cc}} & 0.56 s& $0.831 \pm 0.001$ & $0.909 \pm 0.001$ & $0.734 \pm 0.001$\\
			\midrule
			\multirow{5}{0.15\linewidth}{Weakly-supervised}&\textit{\acrfull{svc}} & 7.8 s & $0.798 \pm 0.028$  &  $0.864 \pm 0.024$ & $0.490 \pm 0.040$\\
			&\textit{\acrfull{mlp}}  & 12.7 s & $0.732 \pm 0.051$  &  $0.813 \pm 0.054$ & $0.017 \pm 0.051$\\
			&\textit{\acrfull{cc} } & 0.83 s & $0.848 \pm 0.005$ & $0.921 \pm 0.003$ & $0.762 \pm 0.005$\\
			&\textit{\acrfull{sae} } & 15.8 s & $\bm{0.872 \pm 0.015}$ & $\bm{0.935 \pm 0.013}$ & $\bm{0.823 \pm 0.017}$\\
			&\textit{\acrfull{cnnsae}} & 132.6 s& $0.866 \pm 0.016$ & $0.931 \pm 0.014$ &$0.798 \pm 0.022$\\			
			\bottomrule
		\end{tabular*}
		
	\end{threeparttable}
\end{table*}

\begin{figure*}[!tb]
    \centering
    \subfloat[\new{FF-AE.}\label{fig:cm_FFAE}]{%
    \includegraphics[width=0.2\linewidth]{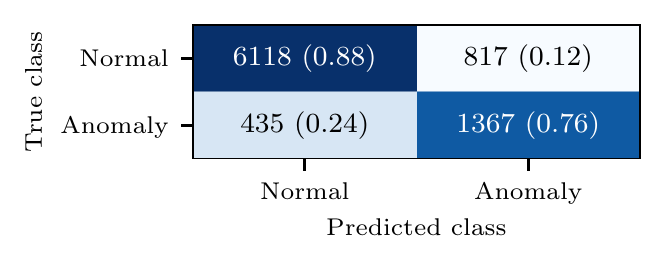}}
    \subfloat[\new{Unsupervised CC.}\label{fig:cm_CC_uns}]{%
    \includegraphics[width=0.2\linewidth]{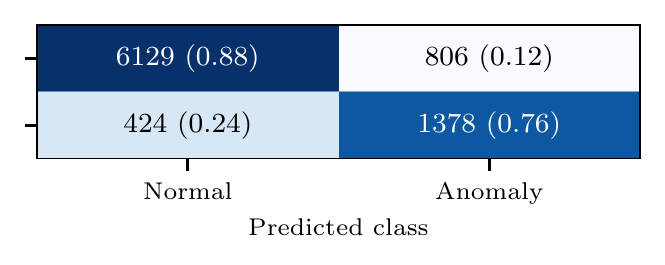}}
    \subfloat[\new{Weakly-supervised CC.}\label{fig:cm_CC}]{%
    \includegraphics[width=0.2\linewidth]{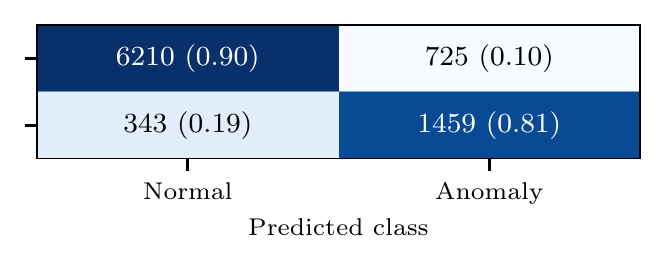}}
    \subfloat[\new{SAE.}\label{fig:cm_SAE}]{%
    \includegraphics[width=0.2\linewidth]{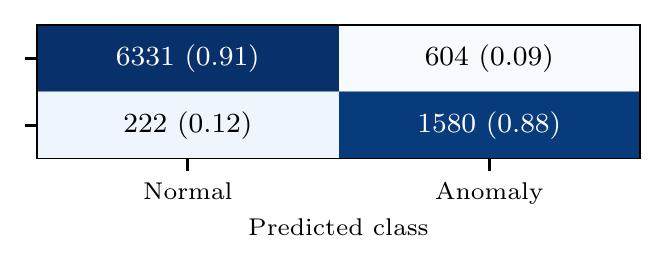}}
    \subfloat[\new{CNN-SAE.}\label{fig:cm_CNNSAE}]{%
    \includegraphics[width=0.2\linewidth]{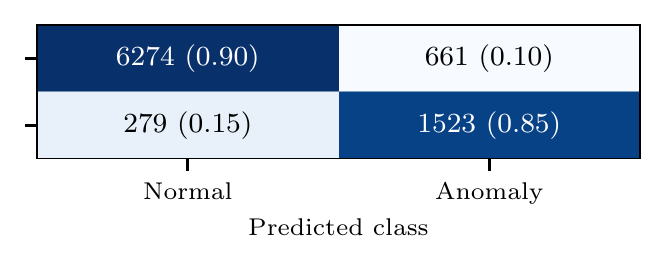}}
    
    \caption{{Confusion matrices of the proposed algorithms trained with Digital Twin data: FF-AE (a), unsupervised CC (b), weakly-supervised CC (c), SAE (d), CNN-SAE (e). The numbers are the absolute detections while the numbers given in parenthesis are the in-class percentages.}}
    \label{fig:cm}
\end{figure*}

\new{First, we report the performance of the proposed unsupervised and weakly-supervised approaches against other state-of-the-art algorithm for anomaly detection, presented in section~\ref{sec:comparative}, applied to a real-world use-case data, introduced in section~\ref{sec:Setup}, by using the Digital Twin to generate the simulation train data. Also, for the weakly-supervised approaches, we analyze how the labeled anomalies used in the training affects the performance.
Then, for completeness, we show the usefulness of the Digital Twin simulation data, by reporting the performance of the anomaly detection algorithms by using only the real-world data.}

\subsection{\new{Results obtained with Digital Twin data simulation}}

\new{The results obtained in our experiments, by using the Digital Twin to generate the training set, are reported in Table~\ref{tab:results}. For the weakly-supervised methods, 10 randomly selected anomalous data samples were included in the training.}


The naive \acrshort{mae} approach of directly comparing Digital Twin  data to real measurement data gives a medium performance of $AP=0.69$. It is worth noting that simple unsupervised clustering-based approaches like KNN and the proposed \acrshort{cc} already give rather good scores of about $AP=0.81$ and $AP=0.83$, respectively. The best performing unsupervised method is the deep learning based \acrshort{ffae} with a slightly better $AP=0.84$ while the training time increases substantially.

All  weakly-supervised algorithms show better performance in all metrics than the unsupervised methods, which is to be expected since additional valuable information on actual anomalies is used during  training.
The best anomaly detection algorithm in our studies is the proposed \acrshort{sae} which reaches an AP score of 0.872, AUC ROC of 0.935 and $F_2$ of 0.823, but with a fairly high variance among our experiments. The larger variance is a direct result of the generation of the training data, since randomly selected samples of normal data are paired with only 10 randomly selected samples from anomalous dataset to produce the input pairs. This produces a large variance in the selected pairs from run to run, which explains the variance in the performance of Siamese approaches.  
The \acrshort{sae} outperforms the weakly-supervised \acrshort{cc} by $2.8\%$, \acrshort{svc} by $19.1\%$ and \acrshort{mlp} by $9.2\%$, in terms of AP score, when trained with the same number of labeled samples.
The \acrshort{cnnsae} also reaches performance levels above the state-of-the-art methods and  comparable to the \acrshort{sae} but requires a much longer training time.

\new{In \figurename~\ref{fig:cm} we show the confusion matrices of the best performing unsupervised algorithm \acrshort{ffae} (\ref{fig:cm_FFAE}) and the proposed methods: unsupervised \acrshort{cc} (\ref{fig:cm_CC_uns}) and weakly-supervised \acrshort{cc} (\ref{fig:cm_CC}), \acrshort{sae} (\ref{fig:cm_SAE}) and \acrshort{cnnsae} (\ref{fig:cm_CNNSAE}).
The false positive rate goes from 12\% to less than 10\% with the change of the training approach from unsupervised to weakly-supervised. And it is is about 9\% for the best performing algorithm in our studies, SAE, which is rather large but just about to be acceptable for the application to the real world. }

\begin{figure}[!tb]
	\centering
	\includegraphics[width=0.95\linewidth]{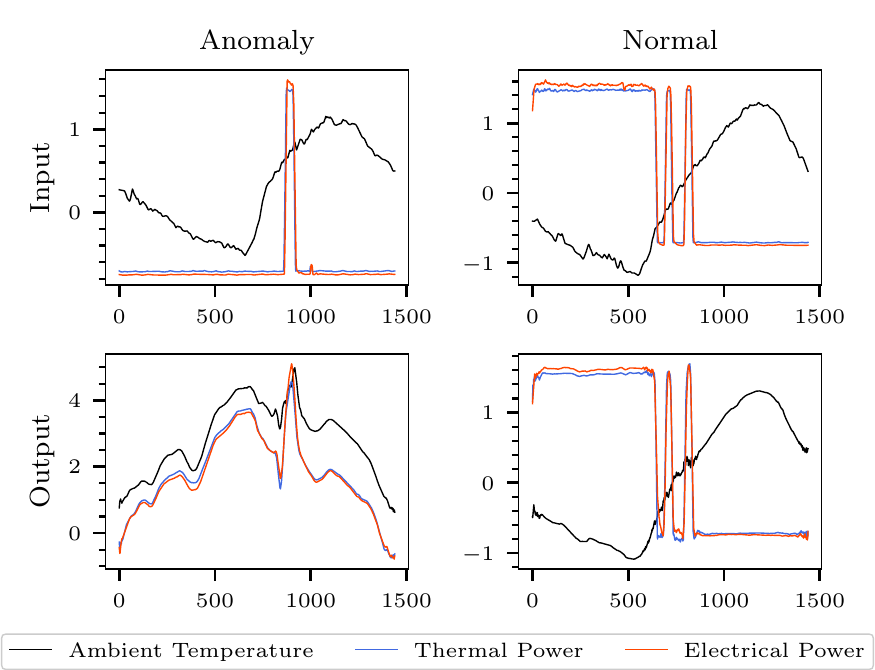}
	\caption{\new{Input, top row, and reconstructed sample, bottom row, of CNN-SAE. The data have been standardized to have zero mean and unitary standard deviation.}}
	\label{fig:cnn_reconstruct}
\end{figure}

\begin{figure}[!tb]
    \centering
    \includegraphics[width=0.925\linewidth]{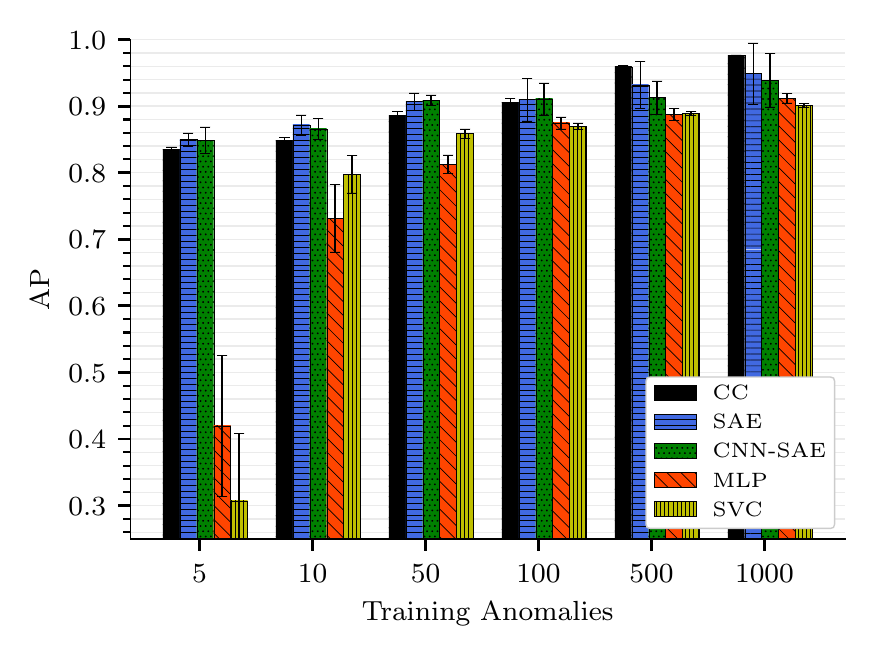}
    \caption{Overall performances of our experiments based on varying the number of labeled anomalies.}
    \label{fig:ano_results}
\end{figure}

\new{An example data from normal and anomaly class and their respective reconstruction, made with the \acrshort{cnnsae}, is shown in \figurename~\ref{fig:cnn_reconstruct}. The upper row shows a normal and anomalous data sequence which is given as input to the model. 
The bottom row shows the reconstruction of the convolutional network. As it can be seen, the initial targets are achieved, since the normal time series is accurately  reconstructed, while the anomalous sample produces an uncorrelated output. }

For the previous results, we used only 10 anomalous data samples in the weakly-supervised setting. In order to get some insights into the dependency on the number of labeled anomalous data samples, we trained the algorithms with differently sized sets of anomalies, $|\mathcal{A}| = \{5, 10, 50, 100, 500, 1000\}$, which corresponds to $0.05\%$, $0.11\%$, $0.57\%$, $1.14\%$, $5.7\%$ and $11.44\%$ of the normal training samples.
As it is expected, there is a general trend that the performance increases with more labeled anomaly samples as observable in \figurename~\ref{fig:ano_results}.
Remarkably, the proposed Siamese approaches already give a rather high AP score above 0.85 with just 5 labeled anomaly samples.
For the largest number of labeled anomalies, the proposed \acrshort{cc} algorithm show the best performance. This is understandable, as in the clustering approach with the penalty term, accordingly to Eq.~\ref{eq:CC}, the information of each anomaly is directly used. However, this is only beneficial as long as  there is no noise in the set of labeled anomalies.  In a real-world  setting with noisy labels, the performance might not improve as much with more labeled data samples. 
The proposed Siamese approaches are always clearly above the state-of-the-art methods. The increase with additional labeled anomaly samples is rather slow which we take as a hint  that the Siamese approaches can be operated rather robustly in a real-world setting with noisy anomaly data.

\subsection{\new{Results obtained with only Real-World data}}

\begin{table*}[!tb]
	\centering
	\caption{\new{Performance comparison of the investigated algorithm trained only with real-world data. The globally best scores are in bold, while the best unsupervised scores are in italic font. The $\Delta_{\%}$ is with respect to the use of Digital Twin data.}}
	\label{tab:results_real}
	\new{
	\begin{tabular*}{\linewidth}{l @{\extracolsep{\fill}} l l l l l}
		\toprule
		\textbf{Training Approach} & \textbf{Algorithm} & \textbf{AP} & $\bm{\Delta_{AP,\%}}$ & \textbf{AUC ROC} & $\bm{\Delta_{ROC,\%}}$\\
		\midrule
		\multirow{7}{0.15\linewidth}{Unsupervised}
		&\textit{\acrfull{if}}      &$0.479 \pm 0.008$  & $-20.6$ &  $0.800 \pm 0.009$ & $-9.9$\\
		&\textit{\acrfull{ocsvm}}   &$0.526 \pm 0.004$  & $-32.4$ &  $0.775 \pm 0.005$ & $-8.1$\\
		&\textit{\acrfull{knn}}     &$0.260 \pm 0.002$ & $-67.8$ &$0.491 \pm 0.002$ & $-45.0$\\
		&\textit{\acrfull{pca}}     & $0.507 \pm 0.002$ & $-2.3$ &$0.614 \pm 0.001$ & $-5.4$\\
		&\textit{\acrfull{lof}}     & $0.295 \pm 0.004$ & $-63.6$ &$0.496 \pm 0.006$ & $-42.9$\\
		&\textit{\acrfull{ffae}}    &$\mathit{0.612 \pm 0.012}$ & $-27.3$ & $\mathit{0.822 \pm 0.012}$ &$-10.1$ \\
		&\textit{\acrfull{cc}}  &$0.468 \pm 0.017$ & $-43.7$ & $0.686 \pm 0.008$ & $-24.5$\\
		\midrule
		\multirow{5}{0.15\linewidth}{Weakly-supervised}&\textit{\acrfull{svc}} & $0.554 \pm 0.085$  &  $-30.6$ & $0.737 \pm 0.036$ & $-14.7$\\
		&\textit{\acrfull{mlp}}  & $0.711 \pm 0.031$  & $-2.9$ & $0.767 \pm 0.026$ & $-5.7$\\
		&\textit{\acrfull{cc}} &$0.584 \pm 0.033$ & $-31.1$ &$0.774 \pm 0.023$ & $-16.0$\\
		&\textit{\acrfull{sae}} & $\bm{0.756 \pm 0.019}$ & $-13.3$ & $\bm{0.876 \pm 0.012}$ & $-6.3$\\
		&\textit{\acrfull{cnnsae}} & $0.589 \pm 0.023$ & $-32.0$ & $0.789 \pm 0.020$ & $-15.3$ \\
		\bottomrule
	\end{tabular*}
    }
\end{table*} 

\new{For the sake of completeness, we show the usefulness of the Digital Twin simulation data by reporting the results obtained by using only the real-world unlabeled data to train the algorithms.
For the weakly-supervised training approach, 10 randomly selected labeled anomalous samples were used during the training. Thus, the anomalous samples were removed from the train-set.
The same networks structure and hyperparameters set of the previous results, were used in those analysis.}

\new{The results obtained by using only the real-world data are reported in Table~\ref{tab:results_real}, in terms of AP, ROC and their difference in percentage with respect to the use of Digital Twin data ($\Delta_{AP,\%}$ and $\Delta_{ROC,\%}$). 
They show the same trend of those with the Digital Twin simulation data, the best performing unsupervised algorithm is the \acrshort{ffae}, and the best algorithm is the weakly-supervised \acrshort{sae}.
But the performance is lower for all the investigated algorithm for both the metrics reported, AP and ROC. 
This is due the fact that the algorithms are not trained with a dataset that reflects only the normal operation modes of the machinery, but the training data is noisy and with some anomalous samples.}


\section{Final Remarks and Conclusion}\label{sec:conclusion}

In this work, we present novel approaches to  multi-variate time-series anomaly detection. We demonstrate the approaches at an application to real-world data from  a facility  monitoring system of  a medium-sized company. We focus on the Combined Heat and Power (CHP) module and use a Digital Twin simulation of the facility to generate normal operation data for training. A small set of labeled anomalies from the actual monitoring system is also available to train the proposed weakly-supervised algorithms. 
 We propose a simple clustering-based approach where the anomaly score includes a penalty term which accounts for the labeled data samples. We also present two approaches realizing Siamese Neural Networks architectures, one taking features derived from the time-series as input and the other directly operating with raw time-series data. These architectures implement autoenconder neural networks, the loss function targets the perfect reconstruction of normal operation data, while anomalous data samples should not be reconstructed well. In order to enhance the discriminatory power of the networks, the latent representations of normal and anomalous samples is forced to have large distances. 
 
 We evaluate many statistical feature types and parameters for the time-series features and observe that simple mean and standard deviation of a time interval of one day, with a stride of one hour, give the best results. 
 We compare the proposed approaches to many state-of-the-art approaches and evaluate a multitude of performance measures for comparison. 
 
 \new{We validate the usefulness of the Digital Twin to generate the normal operation data for training, all the investigated algorithms degrade their performance by a remarkable amount when just the real-world data is used in the training phase.} 
 
All the proposed weakly-supervised algorithms showed better performance that the state-of-the-art approaches according to all performance measures.  The overall best performing algorithm is the \acrlong{sae} operating on time-series features, while the \acrlong{ffae} exhibits best performance of all unsupervised algorithms. 
When varying the number of labeled anomalous samples in the training set, the performance values of the proposed weakly-supervised approaches change mildly and still have very good performance for only five anomalous samples. In contrast, the performance strongly degrades for smaller number of samples in the training set for the \acrlong{mlp} and the \acrlong{svc}.     
 
The false positive and false negative detection rates of the best performing method, \acrshort{sae}, are around 9\% and 12\% in each class, which is acceptable but still rather large for a real-world application.  However,  the current approach constitutes a generic proof of concept for using Siamese Networks for anomaly detection tasks in real-world settings. Multiple options are available to improve the performance for a specific application. Including real-world normal states into the labeled dataset should enhance the performance due to improving the alignment and transfer from purely simulated Digital Twin data  to real measurement data. Instead of just using a fixed predefined threshold for the Anomaly Score,
\new{which is determined in a way to get 25\% anomalies, a more appropriate way should be used where, for example,} the threshold should be learned from the available data and maybe even adjusted with feedback from an expert. 

In the present study, the set of labeled anomalies is assumed to contain no noise, i.e.\ no mislabeled samples. This assumption is not valid in a real-world environment, as labelling errors are to be expected, but also the definition of what constitutes an anomaly is not clear and might change over time. As a consequence, a human-in-the-loop approach with expert feedback is highly desirable where the set of labeled anomalies is re-inspected and possibly extended over time.  
The proposed algorithms are designed to easily incorporate the expert-feedback therein. However, their evaluation is intentionally left for future work. 

\vspace{0.2cm}

\ifCLASSOPTIONcaptionsoff
\newpage
\fi


%


\begin{IEEEbiography}[{\includegraphics[width=1in,height=1.25in,clip,keepaspectratio]{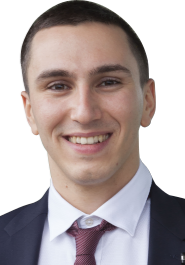}}]{Castellani Andrea}
	was born in Civitanova Marche, Italy, in 1993. 
	He received the B.Sc. and M.Sc. degree (cum laude) in Electronic Engineering respectively in 2016 and 2019 from Universit\'{a}  Politecnica delle Marche (Italy).
	In 2020 he started his Ph.D. perdiod at Uni Bielefeld (Germany) in collaboration with Honda Research Institute (Germany).
	His current research interests include Machine Learning and time-series analysis.
\end{IEEEbiography}
\begin{IEEEbiography}[{\includegraphics[width=1in,height=1.25in,clip,keepaspectratio]{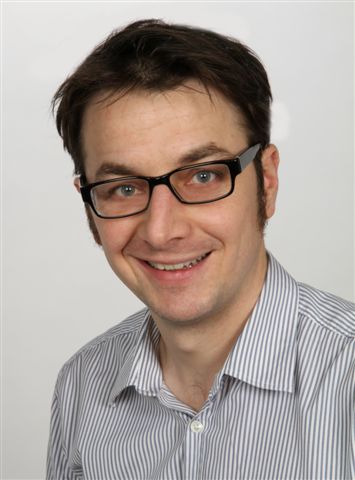}}]{Schmitt Sebastian}
	Sebastian Schmitt was born in 1975 in Germany. He received his PhD in theoretical solid state physics in 2008 from Technical University of Darmstadt, Germany. After a post-doc period from 2009 to 2011 at the Technical University of Dortmund, Germany, he joined the Honda Research Institute Europe in 2012. In his role as Senior Scientist he evaluates recent technologies in optimization, data analytics and machine learning for their use in real-world engineering settings relevant to R\&D departments within Honda. 
\end{IEEEbiography}

\begin{IEEEbiography}[{\includegraphics[width=1in,height=1.25in,clip,keepaspectratio]{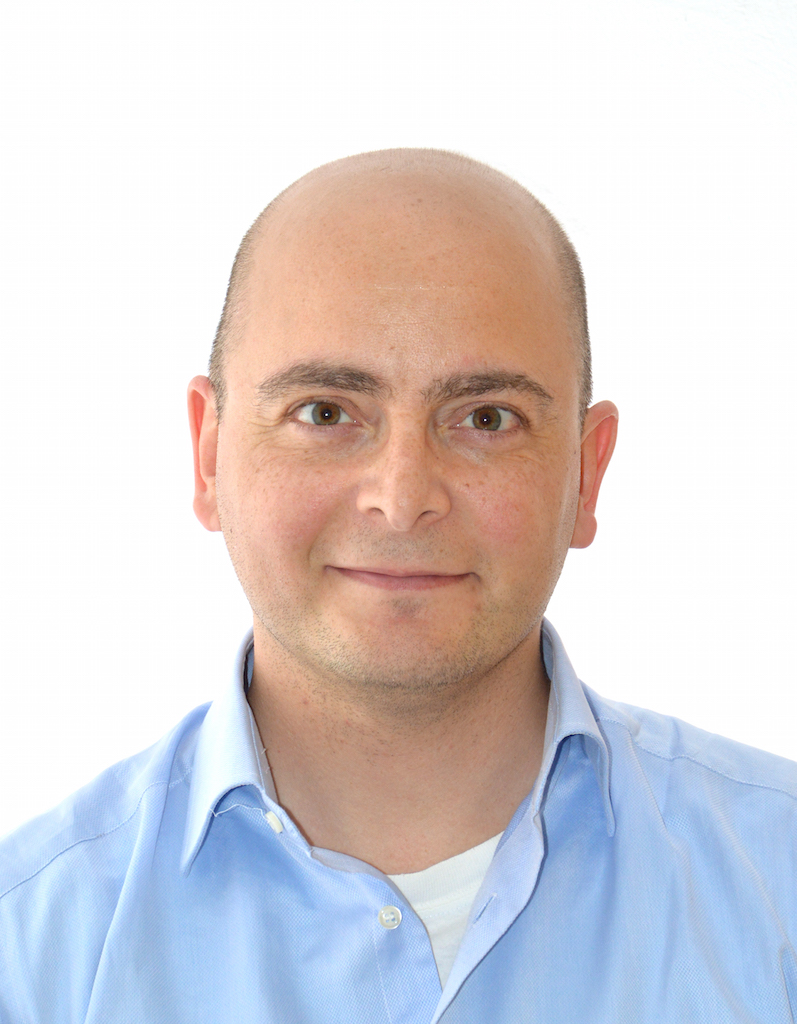}}]{Squartini Stefano}
	 (IEEE Senior Member, IEEE CIS Member) was born in Ancona, Italy, on March 1976. He got the Italian Laurea with honors in electronic engineering from Polytechnic University of Marche, UnivPM, in 2002. He obtained his PhD at the same university in 2005. He joined the Department of Information Engineering as Assistant Professor in Circuit Theory in 2007. He is now Full Professor at UnivPM since 2020. His current research interests are in the area of computational intelligence and digital signal processing, with special focus on audio processing and energy management. He is author and coauthor of more than 210 international scientific papers. He is Associate Editor of the IEEE Transactions on Neural Networks and Learning Systems, IEEE Transactions on Cybernetics and IEEE Transactions on Emerging Topics in Computational Intelligence. He joined the Organizing and the Technical Program Committees of more than 80 International Conferences and Workshops.
\end{IEEEbiography}
\vfill




\end{document}